\documentclass{article}
\usepackage[preprint]{neurips_2024}
\usepackage[utf8]{inputenc}
\usepackage[T1]{fontenc}
\usepackage{hyperref}
\usepackage{url}
\usepackage{booktabs}
\usepackage{amsfonts}
\usepackage{amsmath}
\usepackage{nicefrac}
\usepackage{microtype}
\usepackage{graphicx}
\usepackage{xcolor}
\usepackage{subcaption}
\usepackage{wrapfig}
\usepackage{multirow}

\graphicspath{{./}}

\title{The Boiling Frog Threshold: Criticality and Blindness in\\World Model-Based Anomaly Detection Under Gradual Drift}

\author{
  Zhe Hong\\National University of Singapore
}

\begin{document}

\maketitle

\begin{abstract}
When an RL agent's observations are gradually corrupted, at what drift rate does it ``wake up''---and what determines this boundary? We study world model-based self-monitoring under continuous observation drift across four MuJoCo environments, three detector families (z-score, variance, percentile), and three model capacities. We find that (1)~a \emph{sharp detection threshold} $\varepsilon^*$ exists universally: below it, drift is absorbed as normal variation; above it, detection occurs rapidly. The threshold's \emph{existence} and sigmoid \emph{shape} are invariant across all detector families and model capacities, though its \emph{position} depends on the interaction between detector sensitivity, noise floor structure, and environment dynamics. (2)~Sinusoidal drift is \emph{completely undetectable} by all detector families---including variance and percentile detectors with no temporal smoothing---establishing this as a world model property rather than a detector artifact. (3)~Within each environment, $\varepsilon^*$ follows a power law in detector parameters ($R^2 = 0.89$--$0.97$), but cross-environment prediction fails ($R^2 = 0.45$), revealing that the missing variable is environment-specific dynamics structure $\partial \text{PE}/\partial\varepsilon$. (4)~In fragile environments, agents collapse before any detector can fire (``collapse before awareness''), creating a fundamentally unmonitorable failure mode. Our results reframe $\varepsilon^*$ from an emergent world model property to a three-way interaction between noise floor, detector, and environment dynamics, providing a more defensible and empirically grounded account of self-monitoring boundaries in RL agents.
\end{abstract}

\section{Introduction}

Reinforcement learning agents increasingly rely on learned world models for planning \citep{hafner2023dreamerv3,schrittwieser2020muzero}, but these internal models also offer an underexplored capability: \emph{self-monitoring}. If an agent's world model can predict what will happen next, then systematic prediction failures may signal that something has gone wrong---either in the environment or in the agent's own perception.

Recent work has demonstrated that world model prediction error can detect \emph{abrupt} environmental changes \citep{domberg2025anomaly}. But real-world sensor degradation is rarely abrupt. Cameras fog gradually, LiDAR calibration drifts imperceptibly, and adversarial perturbations may accumulate incrementally. The critical question is: \emph{can an agent detect gradual perceptual corruption, and if so, what determines the boundary between awareness and oblivion?}

We study this question through systematic ablation across four MuJoCo environments, three detector families, multiple hyperparameter configurations, and three model capacities. Our central finding is that the detection boundary is shaped by a \emph{three-way interaction} between the world model's learned noise floor, the detector's sensitivity, and the environment's dynamics structure---not by any single factor alone.

Our contributions are fourfold:
\begin{enumerate}
    \item \textbf{Threshold existence and shape invariance.} We identify a sharp sigmoid detection threshold across three detector families (z-score, variance, percentile) with varied hyperparameters and three model capacities. The threshold's existence and shape are universal; its position is not.
    \item \textbf{Sinusoidal blindness as a world model property.} All detector families---including variance detectors and percentile detectors with no temporal smoothing---are completely blind to periodic drift. This rules out detector-specific artifacts and establishes that the prediction error signal itself lacks drift information under periodic perturbation.
    \item \textbf{Analytical characterization of $\varepsilon^*$.} Within each environment, $\varepsilon^*$ follows a power law in detector parameters ($R^2 = 0.89$--$0.97$), with environment-specific exponents that quantify noise floor structure. Across environments, a global model fails ($R^2 = 0.45$), pinpointing environment dynamics ($\partial\text{PE}/\partial\varepsilon$) as the missing variable.
    \item \textbf{Collapse Before Awareness (CBA).} In fragile environments (Hopper), the agent's policy collapses before any detector fires, creating a regime where drift is lethal but invisible---stable across all detector types and model capacities.
\end{enumerate}

\section{Related Work}

\paragraph{Distribution shift and OOD detection in RL.} Out-of-distribution detection \citep{hendrycks2017baseline} has been extensively studied for neural networks. The concept drift literature \citep{gama2014survey,lu2018learning} addresses distributional shift in data streams using methods like CUSUM \citep{page1954cusum} and Page-Hinkley tests \citep{hinkley1970}. \citet{domberg2025anomaly} apply DreamerV3's world model for anomaly detection under \emph{abrupt} environmental changes. Our work extends this paradigm to \emph{gradual} drift and reveals the threshold phenomenon that abrupt-change studies cannot observe. Critically, we study not whether drift can be detected, but what determines the \emph{boundary} of detectability.

\paragraph{World models.} World models have evolved from simple forward predictors \citep{ha2018world} to sophisticated latent dynamics models \citep{hafner2023dreamerv3,schrittwieser2020muzero}. We use a deliberately simple MLP world model to isolate the threshold phenomenon from architectural complexity, treating the world model as an internal reality representation whose prediction error serves as a self-monitoring signal.

\paragraph{Predictive processing and active inference.} The free energy principle \citep{friston2010free} and ``controlled hallucination'' framework \citep{seth2021being} posit that perception is constrained prediction, with precision weighting governing the balance between prior expectations and sensory evidence. Our noise floor maps onto precision weighting; our $\varepsilon^*$ onto the precision-weighted prediction error threshold; and our sinusoidal blindness onto model evidence optimization absorbing periodic variance. \citet{hobson2014consciousness} propose that dreaming optimizes internal models---our sinusoidal result operationalizes this: the world model ``dreams through'' periodic perturbation by absorbing it as normal variation.

\paragraph{Signal detection theory.} Green and Swets' \citeyearpar{green1966signal} framework for sensitivity-specificity tradeoffs provides the theoretical basis for our detector sensitivity spectrum analysis (Section~\ref{sec:sdt}), where different detectors correspond to different operating points on an environment-specific ROC curve.

\section{Method}

\subsection{Agent and World Model}

We train PPO agents \citep{schulman2017ppo} using Stable-Baselines3 for $10^6$ steps on four MuJoCo-v5 environments: HalfCheetah (obs: 17, act: 6), Hopper (11, 3), Walker2d (17, 6), and Ant (105, 8). For each environment, we train a forward dynamics model $f_\theta(s_t, a_t) \to \hat{s}_{t+1}$: a 3-layer MLP minimizing MSE on transitions collected by the trained policy. Prediction error at each step is $e_t = \|f_\theta(s_t, a_t) - s_{t+1}\|^2$.

To test whether threshold behavior depends on model capacity, we train three variants per environment: \textbf{small} (hidden size 128), \textbf{medium} (512, default), and \textbf{large} (1024).

\subsection{Drift Injection}

Each evaluation episode runs for 1{,}000 steps with 300 unmodified baseline steps. At step $t_0{=}300$, drift is applied to velocity-related observation dimensions:
\begin{equation}
    \tilde{s}_t = s_t + g(\varepsilon, t - t_0) \cdot \mathbf{d}
\end{equation}
where $\mathbf{d}$ is a unit direction vector, $\varepsilon$ controls intensity, and $g$ defines the profile:
\begin{itemize}
    \item \textbf{Linear:} $g = \varepsilon \cdot t$ (monotonically increasing)
    \item \textbf{Sinusoidal:} $g = \varepsilon \cdot \sin(2\pi \cdot 0.01 \cdot t)$ (periodic, zero-mean)
\end{itemize}
We sweep 16 intensities from $\varepsilon{=}10^{-4}$ to $0.5$ with finer resolution near expected thresholds.

\subsection{Detector Family}

We employ three fundamentally different detector families to disentangle threshold effects from detector-specific artifacts:

\paragraph{Doubt Index (DI).} Maintains an exponential moving average of prediction error and flags anomalies via z-score against a pre-drift baseline. Detection triggers when a sliding window of $W$ consecutive steps exceeds threshold $z$. We sweep $z \in \{2.0, 2.5, 3.0, 3.5, 4.0\}$ and $W \in \{1, 3, 5, 10, 20\}$. Note that $W{=}1$ acts as a single-step detector with no temporal integration, yielding high baseline FPR; it is included for completeness but excluded from regression analyses.

\paragraph{Variance detector.} Monitors the variance of prediction error within a sliding window of size $V{=}50$ and flags when variance exceeds $k$ standard deviations above baseline variance, sustained for $C{=}5$ consecutive windows. We sweep $k \in \{2, 3, 5\}$. This detector operates on a fundamentally different signal (second moment rather than first moment of PE).

\paragraph{Percentile detector.} Flags when prediction error exceeds the $p$-th percentile of the baseline PE distribution within a window of $W{=}5$ steps. We test $p \in \{95, 99\}$. Critically, this detector involves \emph{no temporal smoothing}---it compares each observation directly to the baseline distribution.

\subsection{Experimental Protocol}

All conditions use 10 seeds $\times$ 8 episodes per seed = 80 episodes per condition. We report Wilson score 95\% confidence intervals throughout.

\paragraph{Experiment A (Full ablation).} 4 environments $\times$ 9 DI configurations $\times$ 16 intensities $\times$ linear drift = 576 conditions (46{,}080 episodes). Tests shape invariance and extracts $\varepsilon^*$ across hyperparameters.

\paragraph{Experiment B (Sinusoidal drift).} 4 environments $\times$ 6 detector configurations (DI + variance + percentile) $\times$ 16 intensities $\times$ sinusoidal drift. Tests whether blindness is detector-specific.

\paragraph{Experiment B1 (Percentile control).} Percentile detector on sinusoidal drift as a control experiment: if temporal smoothing (EMA) causes sinusoidal blindness, this smoothing-free detector should detect it.

\paragraph{Experiment C (Capacity ablation).} 3 model sizes $\times$ 4 environments $\times$ 16 intensities $\times$ linear drift. Tests whether $\varepsilon^*$ depends on model capacity.

\section{Results}

Table~\ref{tab:overview} summarizes the key findings across environments before detailed analysis.

\begin{table}[t]
\centering
\caption{Overview of experimental findings across four MuJoCo environments.}
\label{tab:overview}
\small
\begin{tabular}{lcccc}
\toprule
Environment & Baseline MSE & $\varepsilon^*$ range (DI) & Sinusoidal & CBA? \\
\midrule
HalfCheetah & 0.163 & 0.0003--0.004 & Blind & No \\
Hopper      & 0.002 & 0.007--0.012 & Blind & \textbf{Yes} \\
Walker2d    & 0.095 & 0.0003--0.003 & Blind & Mild \\
Ant         & 0.025 & 0.0001--0.001 & Blind & No \\
\bottomrule
\end{tabular}
\end{table}

\subsection{Threshold Existence and Shape Invariance}
\label{sec:shape}

\begin{figure}[t]
    \centering
    \includegraphics[width=0.95\textwidth]{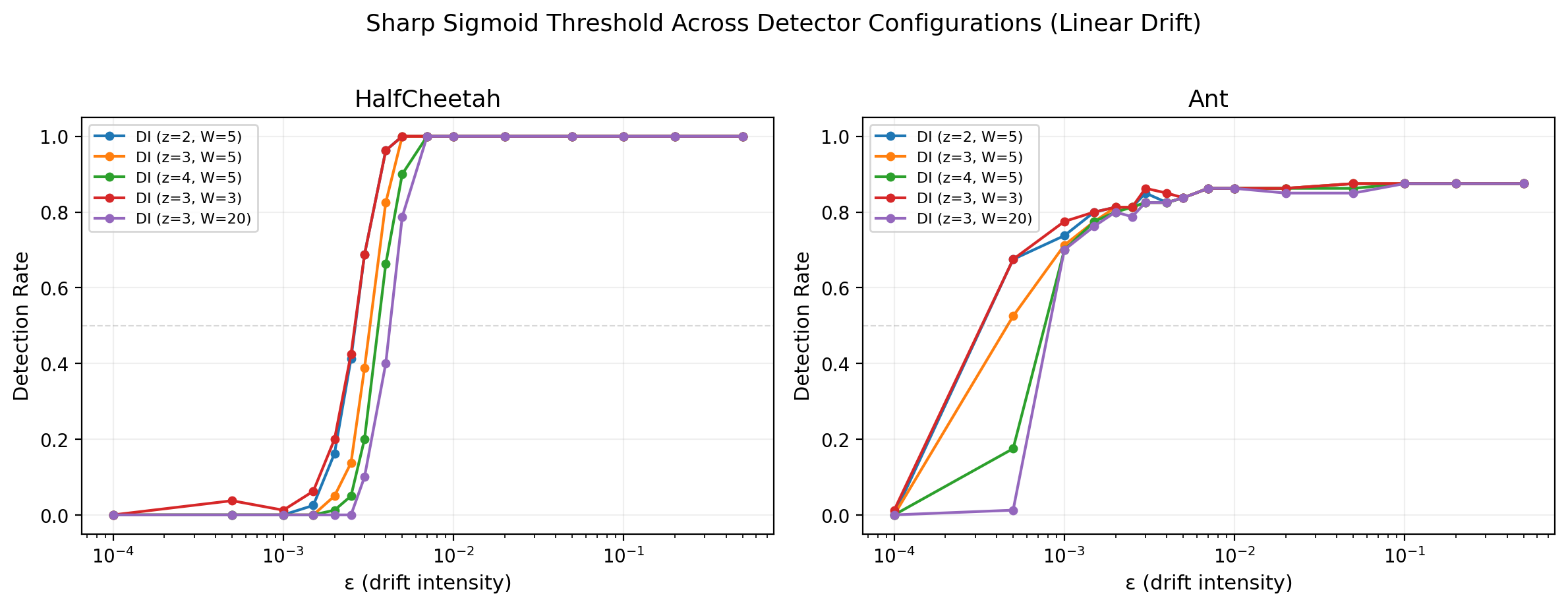}
    \caption{\textbf{Sharp sigmoid threshold across detector configurations (linear drift).} Detection rate vs.\ drift intensity for five Doubt Index configurations in HalfCheetah and Ant. All configurations exhibit the same sigmoid shape; the horizontal position ($\varepsilon^*$) shifts with detector parameters. Full results in Appendix~\ref{app:curves}.}
    \label{fig:sigmoid}
\end{figure}

Across all four environments, all three detector families, and all hyperparameter configurations tested, we observe a consistent pattern: detection rate follows a sharp sigmoid transition from ${\sim}0\%$ to ${\sim}100\%$ as drift intensity increases (Figure~\ref{fig:sigmoid}). This shape invariance is the most robust finding of our study.

The threshold \emph{position} $\varepsilon^*$, however, varies substantially across detectors. Within the Doubt Index family alone, $\varepsilon^*$ ranges from ${\sim}0.0003$ (low $z$, low $W$) to ${\sim}0.004$ (high $z$, high $W$)---nearly an order of magnitude. Across detector families, the range extends further: variance detectors with $k{=}2$ achieve the lowest $\varepsilon^*$ values, while Doubt Index configurations with strict thresholds achieve the highest.

This dissociation between shape invariance and position variability is central to our reframing: the \emph{existence} of a sharp threshold is a property of the world model (all detectors see it), but its \emph{position} is jointly determined by detector sensitivity and the environment's noise floor structure.

\subsection{Sinusoidal Blindness: A Fundamental Limit of PE-Based Monitoring}
\label{sec:sinusoidal}

Our strongest result is that \emph{all} detector families are completely blind to sinusoidal drift across all environments and intensities. Detection rates are indistinguishable from zero, even at $\varepsilon{=}0.5$ where the instantaneous perturbation magnitude far exceeds the noise floor.

This finding is robust to three challenges:

\begin{enumerate}
    \item \textbf{``It's the EMA smoothing.''} The Doubt Index uses an exponential moving average that could, in principle, smooth away oscillatory signals. However, the \emph{variance} detector---which monitors the second moment of PE and uses no such smoothing---is equally blind.
    \item \textbf{``It's temporal averaging.''} The \emph{percentile} detector compares individual PE values directly against the baseline distribution with no temporal aggregation (Experiment B1). It too is completely blind.
    \item \textbf{``It's the detector threshold.''} Different detectors with thresholds spanning a 10$\times$ range all show identical blindness---the signal is not ``just below threshold'' for any detector.
\end{enumerate}

The mechanism is that sinusoidal perturbation oscillates symmetrically around zero: positive and negative deviations cancel over each cycle, so cumulative perturbation never escapes the noise floor. The prediction error signal \emph{itself} contains no drift information---no downstream detector, however sensitive, can extract what is not there. In the language of predictive processing \citep{friston2010free}, the world model performs model evidence optimization by absorbing periodic variance as part of the expected sensory distribution. This parallels \citet{hobson2014consciousness}'s proposal that dreaming serves to optimize internal models---our world model effectively ``dreams through'' periodic perturbation.

A spectral analysis confirms this quantitatively (Figure~\ref{fig:fft} in Appendix~\ref{app:fft}): at $\varepsilon{=}0.01$ in HalfCheetah, linear drift produces PE power $201.6\times$ baseline, while sinusoidal drift produces only $0.8\times$---indistinguishable from the no-drift condition.

\subsection{Collapse Before Awareness}
\label{sec:cba}

In Hopper, we observe a phenomenon stable across all detector types and model capacities: at intermediate drift intensities, the agent's policy physically collapses (the simulated robot falls) \emph{before} any detector accumulates sufficient evidence to trigger. We quantify this via the \emph{survival gap}: $\Delta = T_{\text{collapse}} - T_{\text{detection}}$, where negative values indicate that the agent dies before detection. In Hopper, the collapse rate exceeds 99\% across nearly all drift intensities; at $\varepsilon{=}0.05$, the mean time to collapse is only 25 steps from drift onset while no detector fires at all (Figure~\ref{fig:cba} in Appendix~\ref{app:cba}). This creates four distinct regimes:

\begin{enumerate}
    \item \textbf{Sub-threshold} ($\varepsilon < \varepsilon^*$): drift absorbed, never detected.
    \item \textbf{Detectable} ($\varepsilon$ slightly above $\varepsilon^*$): drift exceeds noise floor; agent survives long enough for detection.
    \item \textbf{Collapse Before Awareness} ($\varepsilon$ moderate): drift destabilizes the policy within ${\sim}30$ steps of onset, but the detector requires ${\sim}50$ steps of post-drift data---the agent dies before waking up.
    \item \textbf{Overwhelming} ($\varepsilon$ very large): even a few steps produce unmistakable signal; detection occurs before collapse.
\end{enumerate}

Note that Hopper exhibits substantial inherent fragility (collapse rate ${\sim}50\%$ even near $\varepsilon{=}0$); the CBA phenomenon specifically refers to drift-\emph{accelerated} collapse that outpaces detection, not baseline instability.

CBA is environment-specific, appearing most strongly in Hopper (fragile single-leg dynamics) and to a lesser extent in Walker2d. HalfCheetah and Ant, with their more stable locomotion, do not exhibit CBA. The practical implication is stark: for fragile agents in safety-critical deployments, there exists a \emph{dangerous blind spot} where perturbations are strong enough to cause catastrophic failure but not strong enough for any detector to fire in time.

\subsection{Detector Sensitivity Spectrum}
\label{sec:sdt}

Having established the qualitative phenomena---threshold existence, sinusoidal blindness, and collapse before awareness---we now turn to quantitative characterization of detector sensitivity across environments.

\begin{figure}[t]
    \centering
    \includegraphics[width=0.95\textwidth]{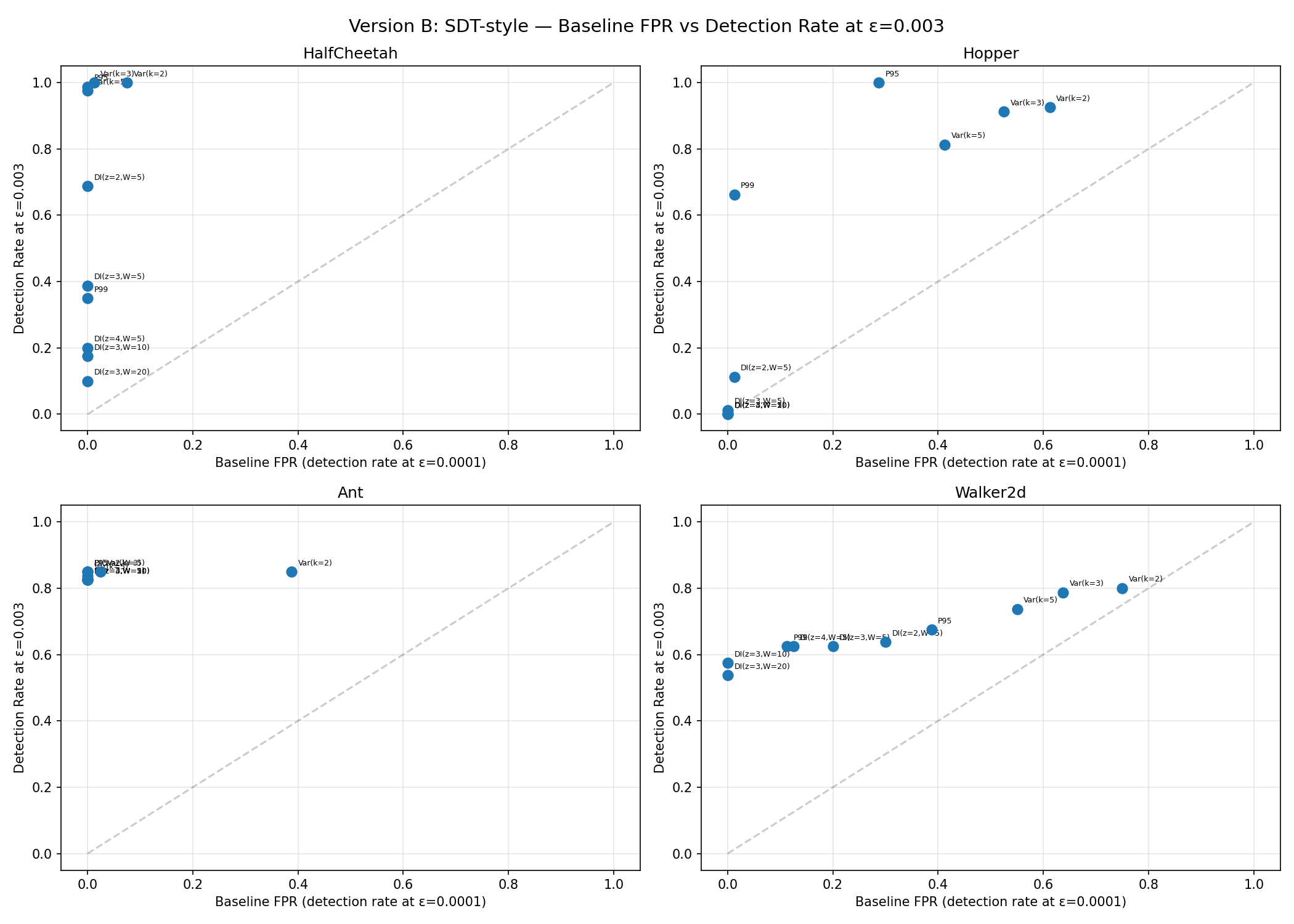}
    \caption{\textbf{Signal detection theory analysis.} Each point represents a detector configuration; $x$-axis is baseline false positive rate (detection rate at $\varepsilon{=}10^{-4}$), $y$-axis is detection rate at reference intensity $\varepsilon{=}0.003$ (chosen as it falls within the transition region for most environments). HalfCheetah and Ant show clear separation (upper-left clustering); Walker2d falls along the diagonal (no detector achieves good sensitivity-specificity separation); Hopper shows a wide spread reflecting the fundamental tradeoff.}
    \label{fig:sdt}
\end{figure}

Figure~\ref{fig:sdt} presents a signal detection theory (SDT) analysis \citep{green1966signal} of all detector configurations. We plot each detector's baseline false positive rate (FPR, measured at $\varepsilon{=}10^{-4} \approx 0$) against its detection rate at a reference drift intensity $\varepsilon{=}0.003$, chosen because it falls within the sigmoid transition region for most environment-detector combinations.

The resulting patterns are environment-specific:
\begin{itemize}
    \item \textbf{HalfCheetah and Ant}: Detector configurations cluster in the upper-left corner (high detection, low FPR), indicating that the drift signal is strong relative to noise and most detectors achieve good separation.
    \item \textbf{Walker2d}: Configurations fall along the diagonal, meaning detection rate increases approximately proportionally with FPR. No detector achieves meaningfully better-than-chance separation at this reference $\varepsilon$---the environment's high baseline noise prevents clean discrimination.
    \item \textbf{Hopper}: Configurations spread widely, reflecting a genuine sensitivity-specificity tradeoff. More sensitive detectors (variance with low $k$) achieve higher detection but at the cost of elevated FPR.
\end{itemize}

Different detector configurations correspond to different operating points on an environment-specific ROC-like curve. The \emph{shape} of this curve---and whether good separation is achievable at all---is determined by the environment's noise floor structure, not the detector family.

\subsection{Analytical Characterization of $\varepsilon^*$}
\label{sec:analytical}

One might expect $\varepsilon^*$ to correlate with baseline prediction error (MSE): environments with noisier world models should have higher thresholds. Surprisingly, this is not the case. Hopper has the lowest baseline MSE ($0.0024$) yet the highest $\varepsilon^*$, while HalfCheetah has the highest MSE ($0.163$) with a relatively low $\varepsilon^*$. The rank ordering by MSE does not match the rank ordering by $\varepsilon^*$ for any detector configuration tested.

\begin{table}[t]
\centering
\caption{Per-environment power law fits: $\varepsilon^* \propto z^\alpha \cdot W^\beta$ (Doubt Index, excluding $W{=}1$). The exponent $\alpha$ quantifies sensitivity to the z-threshold and reflects the noise floor's tail structure.}
\label{tab:powerlaw}
\small
\begin{tabular}{lcccl}
\toprule
Environment & $R^2$ & $\alpha$ (z exponent) & $\beta$ (W exponent) & Interpretation \\
\midrule
HalfCheetah & 0.97 & 0.43 & 0.23 & Clean noise floor; $\varepsilon^*$ insensitive to $z$ \\
Hopper      & 0.95 & 0.44 & 0.45 & Moderate sensitivity \\
Ant         & 0.89 & 1.35 & 0.46 & Heavy-tailed noise; $\varepsilon^*$ highly sensitive to $z$ \\
Walker2d    & 0.78 & 0.61 & 1.00 & High baseline FPR (see Section~\ref{sec:walker2d}) \\
\bottomrule
\end{tabular}
\end{table}

Within each environment, however, $\varepsilon^*$ follows a clean power law in detector parameters (Table~\ref{tab:powerlaw}). For the Doubt Index family (excluding $W{=}1$, which acts as a single-step detector with fundamentally different characteristics), we fit:
\begin{equation}
    \log_{10}(\varepsilon^*) = a + \alpha \cdot \log_{10}(z) + \beta \cdot \log_{10}(W)
\end{equation}

Three of four environments achieve $R^2 > 0.88$, indicating that $\varepsilon^*$ is highly predictable from detector parameters \emph{within} a given environment. The cross-environment global fit, however, yields $R^2 = 0.45$---the missing variable is the environment's dynamics structure, specifically how prediction error responds to drift ($\partial\text{PE}/\partial\varepsilon$).

The \emph{exponents themselves} carry information. The z-exponent $\alpha$ quantifies the noise floor's tail heaviness: Ant's $\alpha{=}1.35$ (heavy-tailed, requiring much stricter $z$ to avoid false positives) versus HalfCheetah's $\alpha{=}0.43$ (light-tailed, $z$ barely matters). This means the noise floor is not merely a scalar (baseline MSE) but has \emph{structure}---different environments produce PE distributions with different shapes, and these shapes determine how detector sensitivity translates to detection capability.

\subsection{Walker2d: A Case Study in High-Noise Environments}
\label{sec:walker2d}

Walker2d exhibits baseline false positive rates of 12--44\% across detector configurations---even at $\varepsilon \approx 0$, detectors frequently fire. This initially appears to undermine the threshold analysis: how can we define $\varepsilon^*$ when the baseline is already noisy?

A FPR-correction analysis reveals that this noise is informative rather than problematic. After correcting for each detector's baseline FPR (subtracting FPR and normalizing), $\varepsilon^*$ \emph{converges} to ${\approx}0.002$ regardless of detector configuration. The corrected $R^2$ drops from 0.78 to 0.36---not because the fit worsened, but because there is almost no remaining variance to explain: all detectors agree on the same threshold once baseline noise is accounted for.

This supports the three-way interaction framework: in high-noise environments, $\varepsilon^*$ is dominated by environmental dynamics rather than detector sensitivity. The apparent detector-dependence of $\varepsilon^*$ in Walker2d is largely a FPR artifact---the \emph{true} threshold is an environment property, masked by detector-specific noise.

\subsection{Capacity Independence}
\label{sec:capacity}

Experiment C tests whether $\varepsilon^*$ is an artifact of model capacity by training small (128), medium (512), and large (1024) hidden-unit world models. As expected, baseline MSE decreases monotonically with capacity (e.g., HalfCheetah: 0.254 $\to$ 0.163 $\to$ 0.145). However, the detection rate curves and $\varepsilon^*$ positions remain essentially unchanged across all three capacities.

This result has a simple explanation: the Doubt Index uses z-score normalization, which divides by the baseline standard deviation. A more accurate model has lower absolute noise but also lower baseline variance, so the z-scores---and thus the detection thresholds---remain the same. $\varepsilon^*$ is invariant to model capacity because the detection mechanism operates on \emph{relative} rather than absolute prediction error.

This rules out the hypothesis that $\varepsilon^*$ reflects model approximation error. The threshold is not ``the model isn't good enough to see small drift''---it is a property of the drift-to-noise \emph{ratio}, which is invariant to uniform scaling of both signal and noise.

\section{Discussion}

\subsection{From Emergence to Interaction}

Our initial hypothesis was that $\varepsilon^*$ is an emergent property of the world model's learned noise floor---a single boundary determined by the model's internal structure. The ablation results require a more nuanced account.

The detection threshold's \emph{existence} and \emph{sharpness} (sigmoid shape) are indeed world model properties: every detector, regardless of family or parameters, sees a sharp transition. Its \emph{position}, however, is jointly determined by three factors:
\begin{enumerate}
    \item \textbf{Noise floor structure}: not just baseline MSE (which fails to predict $\varepsilon^*$ rank order) but the full PE distribution shape, including tail heaviness (captured by the z-exponent $\alpha$).
    \item \textbf{Detector sensitivity}: the operating point on the sensitivity-specificity curve, determined by detector parameters ($z$, $W$, $k$, $p$).
    \item \textbf{Environment dynamics}: how prediction error responds to drift ($\partial\text{PE}/\partial\varepsilon$), which varies across environments and is not predictable from baseline statistics alone.
\end{enumerate}

This reframing is not a retreat---it is a more precise and defensible characterization. What \emph{is} a world model property (threshold existence, shape, sinusoidal blindness) is cleanly separated from what \emph{is not} (threshold position).

\subsection{Predictive Processing Connection}

The three-way interaction framework maps naturally onto predictive processing theory \citep{friston2010free,seth2021being}:

\begin{itemize}
    \item \textbf{Noise floor $\leftrightarrow$ Precision weighting.} The world model's baseline PE distribution determines the ``expected'' level of prediction error---analogous to the precision assigned to prediction errors in predictive coding.
    \item \textbf{$\varepsilon^*$ $\leftrightarrow$ Precision-weighted PE threshold.} Detection occurs when precision-weighted prediction error exceeds a threshold, exactly as in predictive processing accounts of conscious access \citep{dehaene2011experimental}.
    \item \textbf{Detector sensitivity $\leftrightarrow$ Meta-monitoring precision.} Different detectors correspond to different levels of metacognitive precision---a more sensitive detector is analogous to a brain with stronger prefrontal monitoring.
    \item \textbf{Sinusoidal blindness $\leftrightarrow$ Model evidence optimization.} The world model absorbs periodic variance because doing so \emph{increases} model evidence (reduces free energy). This is not a failure but an adaptive response---the model correctly treats zero-mean oscillation as noise rather than signal.
\end{itemize}

\subsection{Implications for Deployed Systems}

Our results yield three practical implications:
\begin{enumerate}
    \item \textbf{Sinusoidal-type perturbations are invisible.} Any drift pattern that oscillates symmetrically around zero will be absorbed by PE-based monitors. Adversaries could exploit this by designing perturbations with zero cumulative drift.
    \item \textbf{CBA creates unmonitorable failure modes.} For fragile agents, some perturbation intensities cause catastrophic failure before any internal monitor can respond. External monitoring (by other agents or supervisory systems) is necessary.
    \item \textbf{Baseline MSE is an unreliable predictor of detection capability.} Practitioners should characterize their environment's $\partial\text{PE}/\partial\varepsilon$ response before deploying PE-based monitors, as model accuracy alone does not predict the detection boundary.
\end{enumerate}

\subsection{Limitations}

Our study is limited to proprioceptive locomotion tasks with MLP world models; generalization to vision-based observations, transformer or RSSM world models, and non-locomotion tasks is unknown. Classical change-point methods (CUSUM, Page-Hinkley) are fundamentally unsuited for monitoring world model PE signals: in our experiments, CUSUM achieves ${>}95\%$ false positive rate even without drift, because these methods assume stationary process noise while the PE signal exhibits inherent non-stationarity from the world model's approximation errors. This underscores the necessity of threshold-based detectors calibrated to the learned noise floor. Drift injection is synthetic (additive, axis-aligned); naturalistic distribution shift may exhibit different threshold behavior. We do not investigate causal attribution---the agent detects \emph{that} something changed but not \emph{what}.

\section{Conclusion}

We have shown that world model-based self-monitoring exhibits a sharp detection threshold whose existence and shape are universal across detector families and model capacities, but whose position reflects a three-way interaction between noise floor structure, detector sensitivity, and environment dynamics. Periodic drift is fundamentally invisible---a world model property, not a detector limitation---because the prediction error signal itself contains no extractable drift information. In fragile environments, collapse before awareness creates a regime where drift is lethal but undetectable by any internal monitor. These findings reframe self-monitoring boundaries in RL agents from simple emergent properties to structured interactions, providing both theoretical insight and practical guidance for deploying self-monitoring agents in non-stationary environments.

Looking ahead, extending these findings to multi-agent settings---where one agent's corrupted perception propagates through interaction---and to vision-based world models (e.g., RSSM, transformer-based) are natural next steps.

\paragraph{Code availability.} Code will be made available upon publication.


\newpage
\appendix

\section{Detection Rate Curves (All Detectors)}
\label{app:curves}

\begin{figure}[h]
    \centering
    \includegraphics[width=0.95\textwidth]{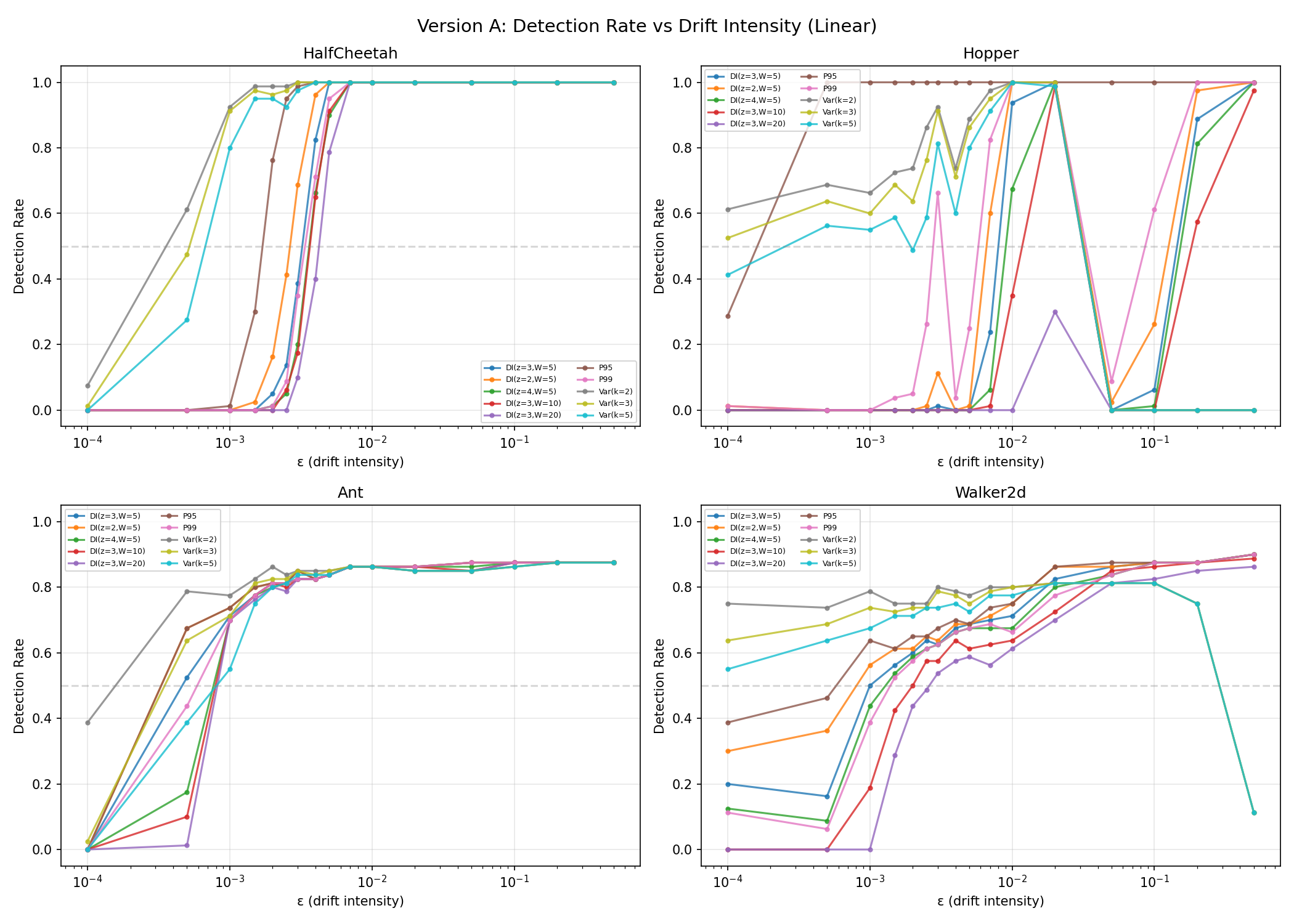}
    \caption{Detection rate vs.\ drift intensity (linear profile) for all detector configurations across four environments. Each curve represents a distinct detector with specific hyperparameters. The sigmoid shape is consistent across all detectors; the horizontal position ($\varepsilon^*$) varies.}
    \label{fig:spectrum_a}
\end{figure}

\section{Per-Environment Regression Details}
\label{app:regression}

Full regression results for the power law model $\log_{10}(\varepsilon^*) = a + \alpha \cdot \log_{10}(z) + \beta \cdot \log_{10}(W)$, including predicted vs.\ actual $\varepsilon^*$ for each detector configuration, are available in the supplementary materials (\texttt{epsilon\_star\_analysis.json}).

\section{Walker2d FPR Correction}
\label{app:walker2d}

For each detector configuration, baseline FPR is measured as the detection rate at $\varepsilon = 10^{-4}$. Corrected detection rates are computed as:
\begin{equation}
    \text{rate}_\text{corrected} = \max\left(0, \frac{\text{rate}_\text{raw} - \text{FPR}}{1 - \text{FPR}}\right)
\end{equation}
After correction, $\varepsilon^*$ converges to ${\approx}0.002$ across all Doubt Index configurations (corrected $R^2 = 0.36$, reflecting convergence rather than poor fit).

\section{CBA Survival Gap Analysis}
\label{app:cba}

\begin{figure}[h]
    \centering
    \includegraphics[width=0.95\textwidth]{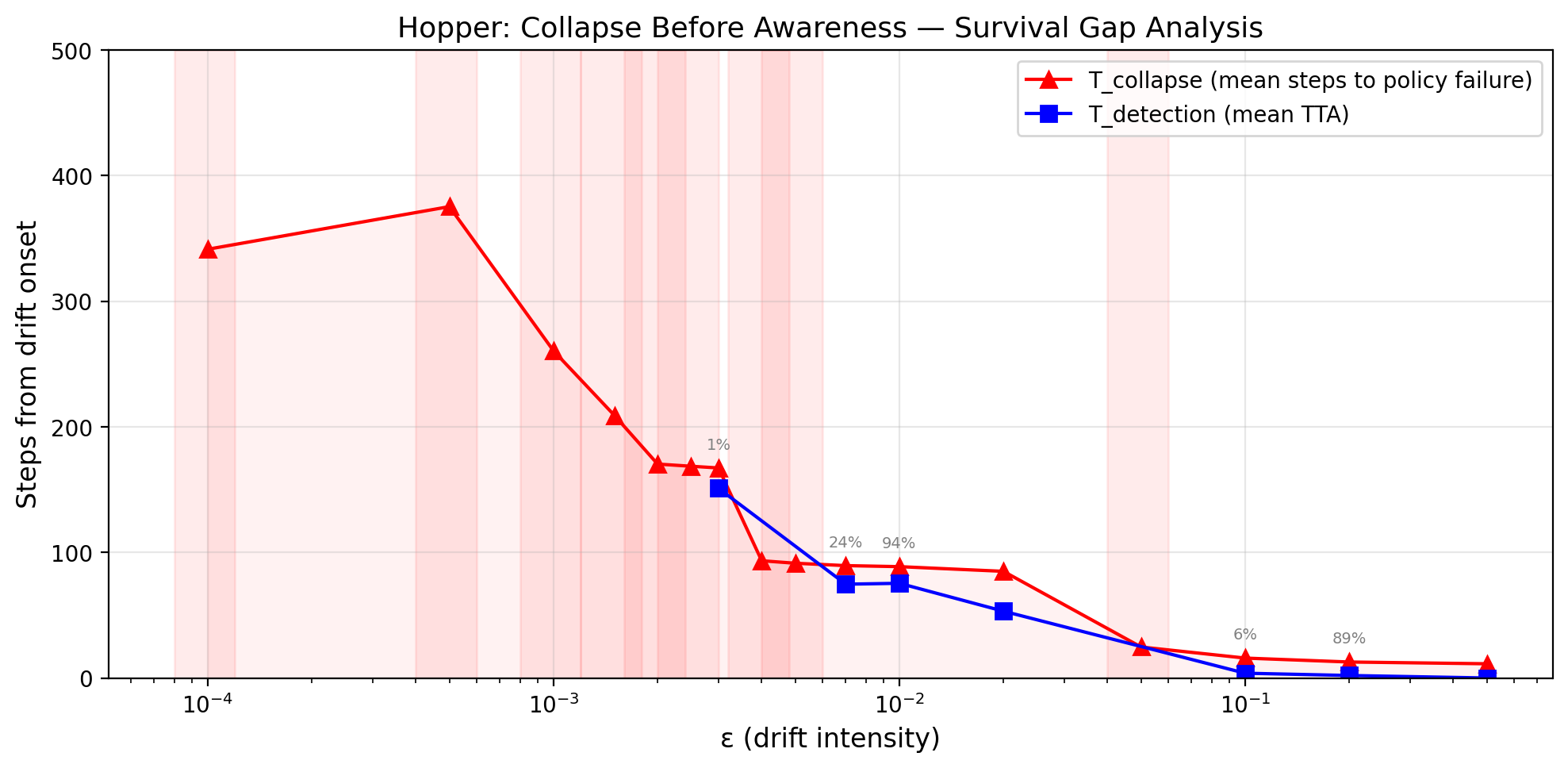}
    \caption{\textbf{Hopper: Collapse Before Awareness analysis.} Red triangles show mean time to policy collapse from drift onset; blue squares show mean time to detection (TTA) for episodes where detection occurs. Hopper collapses at nearly all drift intensities; detection only succeeds when $T_{\text{detection}} < T_{\text{collapse}}$ (high $\varepsilon$). At $\varepsilon{=}0.05$, collapse occurs within 25 steps and no detector fires.}
    \label{fig:cba}
\end{figure}

\section{PE Spectral Analysis}
\label{app:fft}

\begin{figure}[h]
    \centering
    \includegraphics[width=0.95\textwidth]{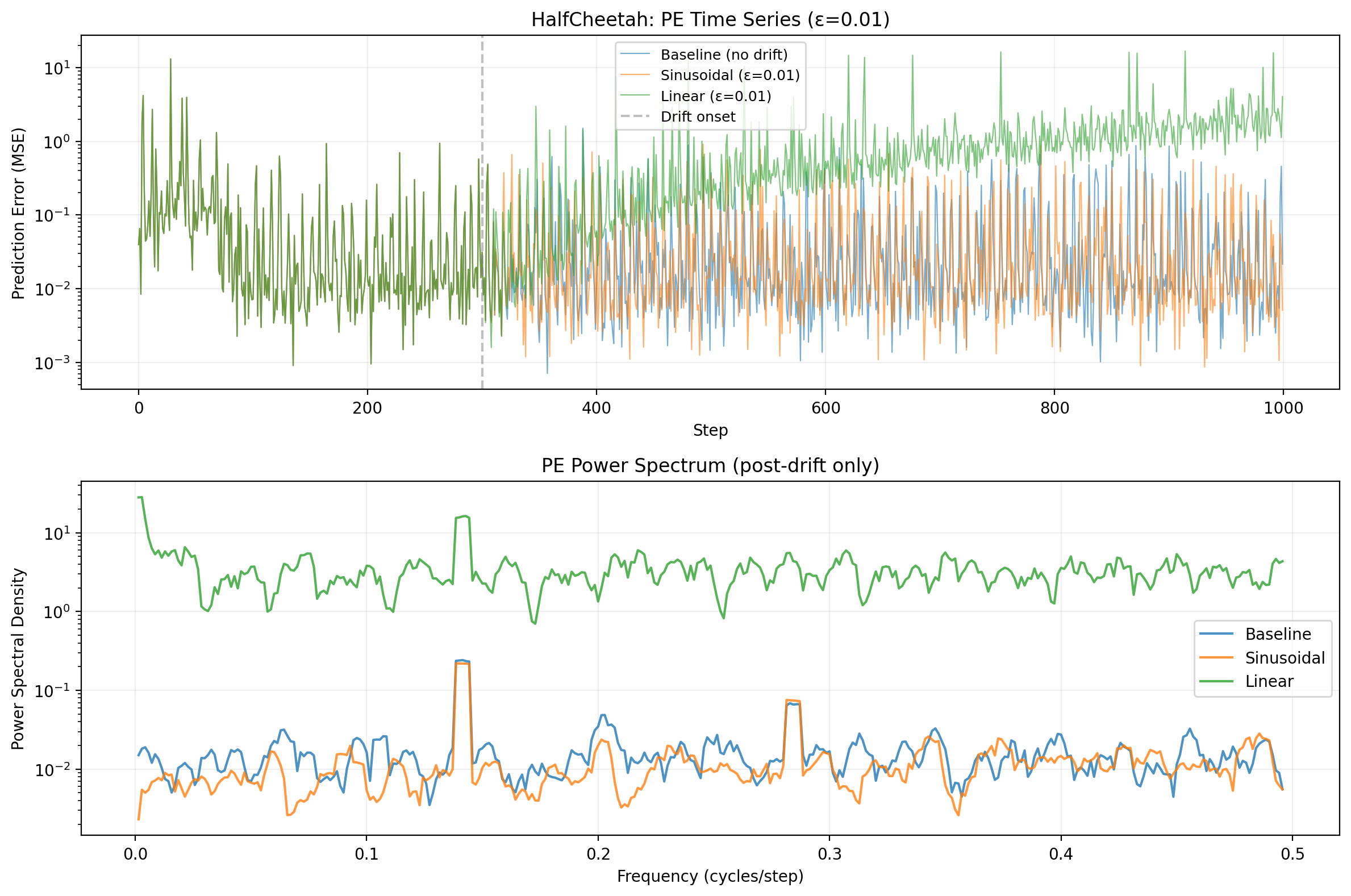}
    \caption{\textbf{Top:} Prediction error time series for three conditions (HalfCheetah, $\varepsilon{=}0.01$). After drift onset (dashed line), linear drift PE diverges while sinusoidal PE remains within baseline range. \textbf{Bottom:} Power spectral density (post-drift). Linear drift exhibits $201.6\times$ baseline power; sinusoidal drift is indistinguishable from baseline ($0.8\times$).}
    \label{fig:fft}
\end{figure}

\end{document}